\pdfoutput=1

\documentclass[11pt]{article}

\usepackage{ACL2023}

\usepackage{times}
\usepackage{array}
\usepackage{latexsym}
\usepackage{graphicx}
\graphicspath{ {./images/} }

\usepackage[T1]{fontenc}

\usepackage[utf8]{inputenc}

\usepackage[noabbrev,capitalize]{cleveref}

\usepackage{microtype}

\usepackage{inconsolata}

\usepackage[normalem]{ulem} 
\usepackage{xcolor}

\title{
Minuteman: Machine and Human Joining Forces in Meeting Summarization \\
}

\author{František Kmječ \\
  Faculty of Mathematics and Physics,\\ 
  Charles University \\
  Czech Republic \\
  \texttt{frantisek.kmjec@gmail.com} \\\And
  Ondřej Bojar \\
  Institute of Formal and Applied Linguistics \\
  Faculty of Mathematics and Physics, Charles \\
  University, Czech Republic \\
  \texttt{bojar@ufal.mff.cuni.cz} \\}

\begin{document}
\maketitle
\begin{abstract}
Many meetings require creating a meeting summary to keep everyone up to date. Creating minutes of sufficient quality is however very cognitively demanding. Although we currently possess capable models for both audio speech recognition (ASR) and summarization, their fully automatic use is still problematic. ASR models frequently commit errors when transcribing named entities while the summarization models tend to hallucinate and misinterpret the transcript. We propose a novel tool -- Minuteman -- to enable efficient semi-automatic meeting minuting. The tool provides a live transcript and a live meeting summary to the users, who can edit them in a collaborative manner, enabling correction of ASR errors and imperfect summary points in real time. The resulting application eases the cognitive load of the notetakers and allows them to easily catch up if they missed a part of the meeting due to absence or a lack of focus. We conduct several tests of the application in varied settings, exploring the worthiness of the concept and the possible user strategies. 
\end{abstract}
\section{Introduction}
When holding meetings, in addition to communicating in real time, it is often necessary to produce an accurate summary of whatever was discussed, what were the major points for and against and what was agreed on. We call the outcome of such a process \textit{a meeting summary} or \textit{meeting minutes}. Such a summary can then go on to be used in subsequent meetings or sent to the participants or those who could not attend but still need to know what happened. 

Meeting summarization is cognitively difficult. This is firstly due to the sheer amount of information the author has to process in real time to be able to write a result of sufficient quality. Secondly, many meetings in non-professional settings do not have a dedicated notetaker and the author has to multitask, on one hand partaking actively in the meeting, and on the other hand writing things down.

Since the coronavirus pandemic began, many meetings have moved to online platforms like Google Meet, JitSi or Zoom. With state-of-the-art technology for ASR and text summarization, it is becoming possible to automate the task. As with most language processing tasks, pre-trained transformer\cite{vaswani2017} language models show the most promise, as shown by \citet{zhang-etal-2022-summn} for summarization in general and \citet{shinde21_automin} for meetings specifically.

\begin{figure*}[t]
    \centering
    \includegraphics[width=\textwidth]{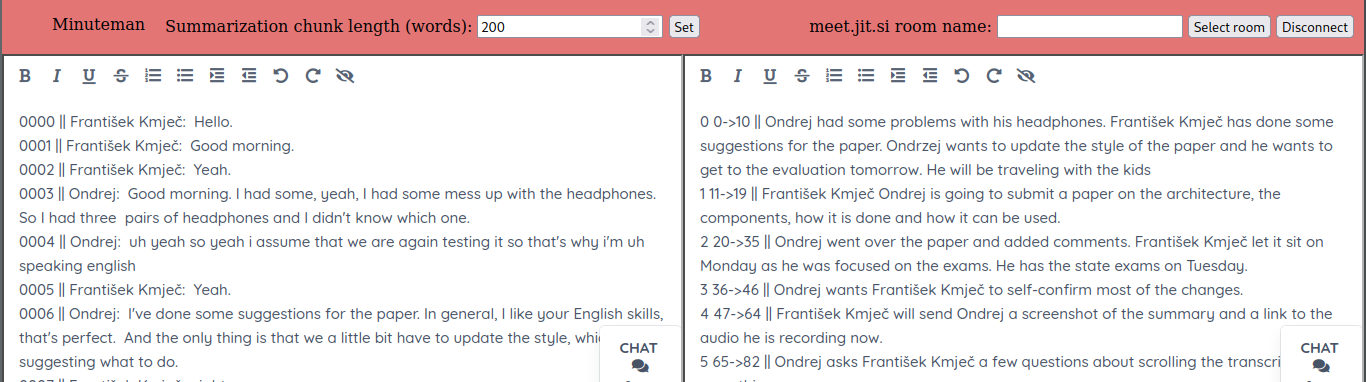}
    \caption{A screenshot of Minuteman during a meeting. Debug mode is enabled so sequence numbers of transcriptions and summaries are shown. The left editor contains the transcript, the right editor contains the summary.}
    \label{fig:minuteman-screenshot}
\end{figure*}

Transformer-based language models have, as of now, a few issues. First of all, they have a limited input size due to the quadratic complexity of the self-attention mechanism, thus reducing the available context.\footnote{Although there are experiments with modifying the attention mechanism to accomodate a longer input, see \citet{beltagy2020longformer}.} Meeting transcripts are often long and will not fit inside a single input window, requiring workarounds. Secondly, current language models are prone to hallucination and can be extremely inaccurate at times, as explored by \cite{hallucination-survey}. But when summarizing meetings, relevance and factuality is key, as many people rely on the output for their work and coordination; thus, mistakes can be costly. Fully automatic solutions already exist,\footnote{see for example \url{sembly.ai} or \url{meetgeek.ai}}, they however do not offer interactivity for the users to control the generated transcript and summary while the meeting is running.

To circumvent the challenges of insufficient summary factuality, coverage and hallucination, we introduce a novel tool, \textit{Minuteman}, to enable effective cooperation between the model and the participants of the meeting. The meeting is recorded and transcribed. The transcript is provided live to the users in an online editor and summarized in real time.
The users can edit the transcript; these edits trigger a new automatic summarization of the respective section, updating the live summary.
The users can also indicate that a particular segment in the transcript is important and trigger its additional automatic summarization.
The live summary is also editable by the users, allowing them to correct or complement the output of the summarization model. The tool is designed in a modular manner, allowing for easy replacement of summarization and transcription models.

\section{Minuteman Tool}
Minuteman is an online application that helps users with meeting minuting. The demo version for testing is available at \url{minuteman.kmjec.cz}. A screenshot of the application user interface is shown in \cref{fig:minuteman-screenshot}.

Upon entering a Jitsi\footnote{\url{meet.jit.si}} room name, Minuteman connects to the meeting as an additional participant to record all participants' audio tracks. A~live transcript is then generated from the conversation of the meeting participants, relying on speech recognition (\cref{sec:asr} below) and summarization (\cref{sec:summ}).

The summary creation works automatically in an iterative manner; when enough new utterances have been appended, a new summary point is created to represent them. The density of the summary can be controlled by selecting the chunk length in words in the top bar. This selection only affects the newly generated summary points.

\begin{figure}[t]
    \centering\includegraphics[width=8cm]{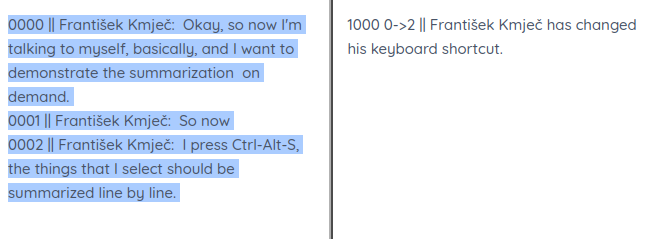}
    \caption{Segment selection and on-demand summarization}
    \label{fig:summary-on-demand}
\end{figure}

It is also possible to select a portion of the transcript and trigger its summarization manually by pressing \texttt{Ctrl + Alt + S}. This additional summary point is appended at the end of the minutes at the point when it was created, see \cref{fig:summary-on-demand} for an example.

The transcription and summarization models can produce errors. Minuteman thus allows the meeting participants to correct them using two  shared editors. Both the transcript and the generated summaries are editable by anyone, with changes in transcript being reflected in the generated summaries. However, if a summary point is edited, it is then frozen and never updated by Minuteman again, to prevent the model from overwriting users' corrections.

\subsection{Implementation and Architecture}
\begin{figure*}
    \centering
    \includegraphics[width=\textwidth]{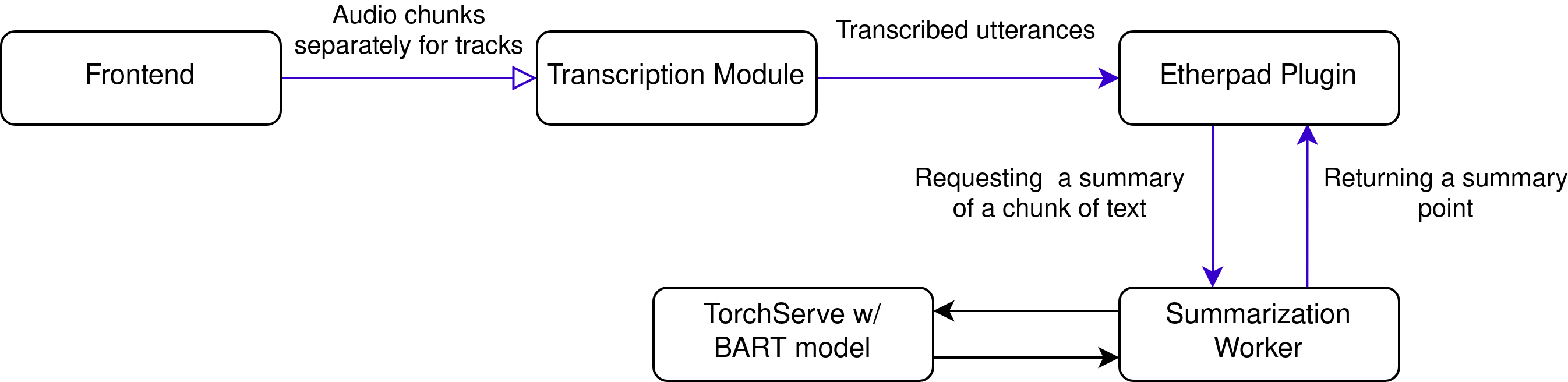}
    \caption{Application architecture. The blue markers represent data being sent over \texttt{RabbitMQ}.}
    \label{fig:app-architecture}
\end{figure*}
Minuteman consists of four main components: the frontend user interface and sound recording, the transcription module, the backend of the editor and the summarization module. These components communicate together over a \texttt{RabbitMQ}\footnote{\url{rabbitmq.com}} message queue, allowing for easy interchangeability. The components are containerized using \texttt{Docker} and built and run using \texttt{docker-compose}.

\subsection{Frontend and Sound Processing}
The frontend is responsible for providing the user with two Etherpad editors and the control bar, and for recording the audio. 
When Minuteman connects to the meeting, it sets up audio recording for each user track separately, therefore it does not need diarization to distinguish different users. The audio is converted to 16KHz and sent to the backend in one-second long chunks for transcription. The audio recording itself happens in in a separate Javascript thread, meaning it does not block the UI processing code.

The recorded chunks are bound to an identifier of the track and session and sent to a Python Flask\footnote{\url{flask.palletsprojects.com/en/2.3.x/}} API, which appends them to a processing queue provided by \texttt{RabbitMQ}, ensuring good ordering. The chunks are then picked up by the transcription module and processed.

\subsection{Transcription Module}
\label{sec:asr}

The transcription module collects recorded audio chunks from a queue and processes them, producing transcript utterances. It is a Python script connected to \texttt{RabbitMQ} via the \texttt{pika}\footnote{\url{github.com/pika/pika}} library. It uses a \texttt{Whisper} ASR model by \citet{radford2022robust} for transcription.\footnote{namely the \texttt{faster-whisper} implementation from \url{github.com/guillaumekln/faster-whisper}} As \texttt{Whisper} is not purposefully built for live transcription but for transcribing already recorded audio files, the live ASR requires a mechanism for splitting the audio tracks into utterances and transcribing those as a whole.

We keep a buffer of audio data for each track, and we ensure that the all chunks in the buffer always contain speech. When a new audio chunk comes, it is first checked for speech presence using Silero voice activity detector (VAD) by \citet{silerovad}. If speech is detected, the chunk is appended to the buffer. If no speech is detected, we assume that the utterance in the buffer is finished and we send the buffer contents to \texttt{Whisper} for transcription. We then flush the buffer contents. The transcribed utterance is sent to the editor backend over \texttt{RabbitMQ}. This setup implicitly means that utterances are sorted by their end times. We chose this approach due to simplicity, however, there may be a need for more complex utterance ordering in the future. A known limitation is that we currently do not support more speakers in the same audio channel.

\subsection{Editor Backend}
To allow for collaboration of multiple users as well as interaction with the summarization model, we use the Etherpad editor. We implement the tool backend as an Etherpad plugin written in Javascript with dependencies managed through \texttt{npm}.\footnote{\url{npmjs.com}}

The first responsibility of the backend is handling the appending of utterances from the transcription module to the transcript editor pad. Each utterance is given a sequence number, which is bound to the utterance line contents using Etherpad attributes. This is done to be able to refer to transcript sections even when they are being edited by the users.

Secondly, the backend handles the summarization of the transcript. As utterances are received from the transcription module, the plugin keeps track of how many unsummarized words are present. When the number of unsummarized words reaches a threshold, the section of transcript to be processed is extracted and sent to the summarization module. The starting and ending utterance sequence numbers are saved. An impromptu summarization message is placed in the summary pad and once a response comes back from the summary module, it is replaced by the generated summary. The same process is repeated when a user requests a summary of a certain selected segment.

An important part of the editor backend code is the extraction of transcript segments. To ensure robustness with respect to user edits, we work at the level of single utterances (lines). The extraction mechanism is given a sequence number of the starting and ending utterance; it then iterates over the whole transcript. When it finds the starting utterance or an utterance with a higher sequence number, it starts recording the transcript segment. When it reaches the ending utterance or an utterance with a higher sequence number, it stops the recording and returns the recorded segment for summarization. That way, the process behaves robustly and predictably even in a collaborative environment.

Lastly, the backend ensures that when the transcript changes (perhaps due to a user edit that corrects a badly transcribed utterance), the summaries generated from the affected segments are updated. On every edit, the already-summarized segments are extracted and compared to their past form; if we find a difference, the segment is summarized again and the corresponding summary is updated. However, if the summary point was already edited by a user, it is not overridden to preserve the user inputs.

\subsection{Summarization Module}
\label{sec:summ}

The summarization module is a Python program listening on \texttt{RabbitMQ} for incoming transcript chunks. It requests the summaries from a BART model by \citet{lewis-etal-2020-bart} finetuned on XSum \cite{narayan-etal-2018-dont} and SAMSum \cite{gliwa-etal-2019-samsum} datasets. The model is available from HuggingFace.\footnote{\url{huggingface.co/lidiya/bart-large-xsum-samsum}} We elected to use BART because it provided one of the best performances at the Automin 2021 competition \cite{ghosal21_automin}, with a successfull team \cite{shinde21_automin} using it together with several preprocessing steps. We take up the same preprocessing process, including the clearing of stopwords and removing unnecessary filler words. To enable simple interchange of models for newer ones, we run the model in a TorchServe\footnote{\url{pytorch.org/serve/}} backend.

\section{User Testing}
We conducted several tests of the tool between the authors and together with a group of network administrators from a local high school, using their work meeting as a testing ground for meetings with multiple active participants. We exploited the fact that their meetings contain a lot of named entities and technical wording, allowing us to test the ASR model to the limit. Based on the results, we formed a qualitative assessment of the tool usability and possible workflows. All the participants of our experiments were briefed and consented to their recordings being used in the evaluation. 

\subsection{Suggested Workflow}
An efficient workflow is reliant on having multiple available participants in the meeting to supervise the transcript and summary points; we found it difficult to keep track of what was happening in the transcript and in the summary in only two people, as constant activity is required of both of the participants. However, the summary was of high quality, capturing the contents of the meeting well, and if the group of two users needs to produce a summary anyway, the tool definitely helps. 

In a larger group of users, only several of them are usually vocally active. The rest can then contribute to correcting the transcript and the generated summary. Upon testing with the administrator group, we found that the workflow of transcript correction is effective, allowing everyone to correct named entities misidentified by the model. At the same time, we observed that the overall summary quality was perceptibly lower. This could be due to a number of causes including lower microphone quality, different non-native accents, less well-arranged transcript due to more participants or the fact that the summary model was finetuned on short non-technical conversations. We give a detailed overview of the errors encountered below as well as suggestions for future work to counteract them.

\subsection{Error Analysis}
We found that most errors were committed by the ASR model upon transcribing named entities. This was expected; many of the topics discussed in the test meetings required sufficient domain knowledge or were in different languages. Examples of transcription errors are listed in \cref{tab:whisper_example_errors}. These could probably be largely counteracted by using a more powerful version of the \texttt{Whisper} model; while testing, we resorted to the \texttt{small.en} variant due to speed and hardware constraints. Also, many of the errors originate in the non-native English of the meeting participants with imperfect pronunciation and in bad quality of the participants' microphones. Upon inspecting the transcript, we updated the model to \texttt{medium.en}, increasing the transcript quality without sacrificing much speed.

\begin{table}[t]
    \centering
    \begin{tabular}{ m{10em} | m{8em} }
         Example & Error explanation \\
         \hline \hline
         Vojta:  a different DHCP server named \textbf{care} so we can try it, I've never used it. & The discussed DHCP server is called Kea, not `care'. \\
         \hline
         Fanda:  like, adapt this towards \textbf{check} summarization, like, you just, like, one thing is swapping for \textbf{check} whisper, that's easy, and one thing is just, like, uploading a new model to\dots{}  & The \texttt{Whisper} model misinterpreted bad pronunciation and did not recognize the word `Czech' in context.
    \end{tabular}
    \caption{Examples of errors committed by the ASR model}
    \label{tab:whisper_example_errors}
\end{table}

As for summarization model errors, from our experiments, we conclude that the quality of the generated output is highly dependent on the quality and coherence of the provided transcript. We divide the committed errors into three main categories. Examples are provided in the list below: \\
\begin{itemize}
    \item \textbf{Overgeneralization:} \texttt{PARTICIPANT1 and PARTICIPANT2 discuss the implementation of a text editor.} is a true statement for our test meeting, but it does not convey any important information that would be worth writing down, since the entirety of it was devoted to improving the editor.
    \item \textbf{Swapping or misinterpreting the actors of an action:} \texttt{PARTICIPANT1 wants PARTICIPANT2 to finish the machinery before the end of this month so that if she switches the cables, she can just note it down and some scripts will fix it for him.} It is noted that PARTICIPANT1 wants PARTICIPANT2 to do something, but this is never mentioned in the transcript.
    \item \textbf{Errors due to lacking context:} \texttt{PARTICIPANT1 needs to refer to some parts of the transcript for the minutes to get summarized. PARTICIPANT2 will double check the deadline for the bachelor thesis.} In the transcript, the checked date was supposed to be the deadline of paper submissions, not for the thesis, but it was discussed in the same context as the bachelor thesis. From a longer context window, the error could be deduced by the model and avoided.
\end{itemize}
Overall, it can be stated that the generated summary is good at capturing the main point of contention for a summary segment, but it very often fails on determining who is the subject of an action and who is an object; much user cooperation is needed in that regard. The generated summary also does not necessarily correspond to a predetermined meeting agenda; it can therefore be difficult for the users to manipulate the model to focus on the content that is important to them. This is however natural, as the model cannot know the agenda in advance.

\subsection{Feedback From Users}
The testers reported that they appreciated the possibility of catching up with the meeting even on getting a quick pause. They did not yet feel comfortable trusting the tool for summarizing the whole meeting, noting the differing styles of a normal summary that mostly focuses on agreed-upon conclusions stemming from a previous agenda and the generated summary. Overall, they found the tool helpful for catching up with the meeting when they were interested to know what each participant had to say about a given point.

\section{Conclusion}
We demonstrated a novel modular tool for interactive summarization, implementing a promising user interface concept. We conducted several qualitative evaluations of the tool outputs and collected feedback from the users. From the user feedback, we conclude the tool demonstrates a worthy concept, although a lot of improvement is needed for the summaries to be reliable and trusted by the users completely. 

\subsection{Future work}
We divide possible improvements into three categories:

\paragraph{Summarization Models}
We presume using larger models like Llama introduced by \citet{touvron2023llama} would be helpful for getting more relevant summaries. A possible improvement would be finetuning these for summarization and replacing BART. Also, the summarization module could be modified to request summaries from ChatGPT API,\footnote{\url{openai.com/blog/chatgpt}} allowing efficient cooperation with a powerful model that is currently very popular with users. Prompt engineering would then be required to support summarization.

\paragraph{User Interface}
Currently, it can be difficult for the user to find where the summary points refer to in the transcript. Adding color coding to the summaries and the transcript segments would greatly improve the orientation in the transcript-summary correspondence. Inspiration could be drafted from the ALIGNMEET tool for meeting evaluation by \citet{polak-etal-2022-alignmeet}, which also uses color coding for this purpose.

\paragraph{Underlying Meeting Platform}
Currently, a large limitation is the restriction to the Jitsi Meet platform. A possible improvement is to rewrite the interface to allow the user to connect to Google Meet, Zoom etc. It would also be worthwhile to prepend the transcription pipeline with diarization and record in-person meetings, allowing the tool to be used in offline settings.

\section{Limitations}
We note that our assessment of the result quality is only qualitative and of a limited sample size, as we did not have the means to conduct a larger quantitative testing effort. Testing was carried out in English by non-native English speakers, therefore the quality of the results can be influenced by non-natural word orders and phrases taken over from their mother tongue (Czech). 
\bibliography{anthology,custom}
\bibliographystyle{acl_natbib}



\end{document}